\documentclass[runningheads]{llncs}
\usepackage[T1]{fontenc}
\usepackage{graphicx}
\usepackage{amsmath} 
\usepackage{algorithm}
\usepackage{algpseudocode}
\usepackage{subfig}
\usepackage{multirow} 
\usepackage{graphicx}
\usepackage{tikz}
\usetikzlibrary{positioning, shapes, arrows}
\begin{document}

\title{EBA-AI: Ethics-Guided Bias-Aware AI for Efficient Underwater Image Enhancement and Coral Reef Monitoring}

\author{Lyes Saad Saoud\inst{1} \and
Irfan Hussain\inst{1}\thanks{Corresponding author: \email{irfan.hussain@ku.ac.ae}}}


\institute{
Khalifa University Center for Autonomous Robotic Systems,\\
Khalifa University, Abu Dhabi, United Arab Emirates\\
}

\maketitle

\begin{abstract}
Underwater image enhancement is vital for marine conservation, particularly coral reef monitoring. However, AI-based enhancement models often face dataset bias, high computational costs, and lack of transparency, leading to potential misinterpretations. This paper introduces EBA-AI, an ethics-guided bias-aware AI framework to address these challenges.
EBA-AI leverages CLIP embeddings to detect and mitigate dataset bias, ensuring balanced representation across varied underwater environments. It also integrates adaptive processing to optimize energy efficiency, significantly reducing GPU usage while maintaining competitive enhancement quality.
Experiments on LSUI400, Ocean\_ex, and UIEB100 show that while PSNR drops by a controlled 1.0 dB, computational savings enable real-time feasibility for large-scale marine monitoring. Additionally, uncertainty estimation and explainability techniques enhance trust in AI-driven environmental decisions.
Comparisons with Cycle-GAN, FunIEGAN, RAUNE-Net, WaterNet, UGAN, PUGAN, and UT-UIE validate EBA-AI’s effectiveness in balancing efficiency, fairness, and interpretability in underwater image processing. By addressing key limitations of AI-driven enhancement, this work contributes to sustainable, bias-aware, and computationally efficient marine conservation efforts.  For interactive visualizations,
animations, source code, and access to the preprint, visit https://lyessaadsaoud.github.io/EBA-AI/.

\end{abstract}

\section{Introduction}

Artificial intelligence (AI) is revolutionizing marine conservation by enabling large-scale coral reef monitoring and climate impact assessment. AI-powered underwater image enhancement and dehazing improve visibility, facilitating biodiversity analysis and environmental evaluation. However, current models face three critical challenges: dataset bias, high computational demands, and lack of transparency.

\textbf{Dataset bias} undermines generalizability. Many enhancement models are trained on tropical reef images, limiting adaptability to diverse ecosystems. Variations in species, water temperature, and turbidity often degrade performance, particularly for cold-water and deep-sea reefs \cite{mukonza2022micro,mcclanahan2024complex,minai2024evaluating,villon2022confronting}. Moreover, most datasets comprise clear, well-lit images, reducing robustness in turbid or low-visibility scenarios \cite{liu2025unsupervised,williams2019leveraging}.

\begin{figure*}[t]
    \centering
    \includegraphics[width=1\textwidth]{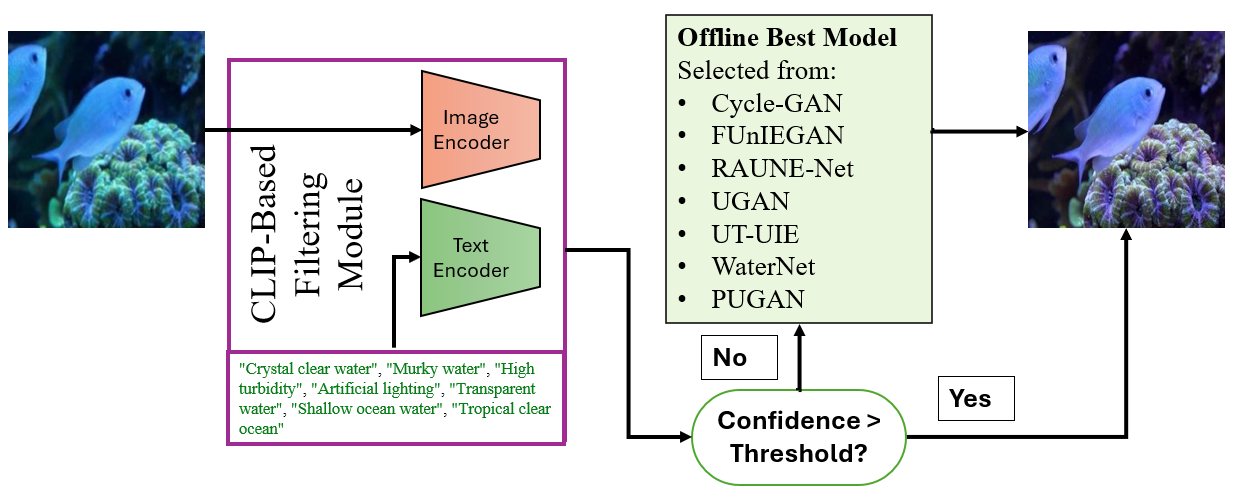}
    \caption{Illustration of the EBA-AI framework. Input images are filtered by CLIP-based clarity scoring; low-quality frames are enhanced using the best offline model. Enhanced outputs are compared to ground truth (GT) for evaluation.}
    \label{fig:main_figure}
\end{figure*}

\textbf{Computational inefficiency} is another concern. Deep models require significant energy, contributing to high carbon emissions. Training a single network can emit CO$_2$ levels comparable to multiple cars over their lifetime \cite{selvan2023operating,xu2023energy,desislavov2023trends}. Energy-efficient neural architecture search (NAS) and quantization techniques are increasingly vital for reducing AI’s environmental impact \cite{bakhtiarifard2024ecnas}.

\textbf{Transparency limitations} hinder trust and validation. Many state-of-the-art models function as black boxes, making it difficult to interpret predictions or verify image reliability. Inaccurate coral health estimates risk misdirected conservation actions. Explainable AI (XAI) is essential for interpretability and confidence in AI-assisted decision-making \cite{welle2017estimating,hoadley2024bio,gonzalez2023integrating}.

To address these issues, we introduce {EBA-AI} (Ethics-Guided Bias-Aware AI), a novel framework that improves reliability, energy efficiency, and interpretability in coral reef monitoring. Its contributions include:

\begin{itemize}
\item \textbf{Bias Mitigation}: CLIP-based embeddings identify and reduce dataset bias for broader ecological coverage.
\item \textbf{Energy-Efficient AI}: Adaptive inference minimizes GPU usage and carbon footprint.
\item \textbf{Trust and Transparency}: XAI and uncertainty estimation enhance interpretability and decision support.
\end{itemize}

EBA-AI offers an ethical, sustainable approach for underwater image processing. Figure~\ref{fig:main_figure} outlines its key modules, from clarity assessment to model selection and transparent evaluation.

\section{Background \& Related Work}

\subsection{Bias Challenges in AI-Driven Coral Reef Classification}

AI has become integral to coral reef classification, enabling large-scale monitoring and conservation. However, dataset disparities and image processing biases often lead to misleading reef health assessments and misinformed conservation strategies. Many models are trained on clear-water reef datasets, making them unreliable in turbid or deep-sea environments \cite{chegoonian2017comprehensive,jackett2023benthic,lam2018acute}. This overfitting to specific conditions limits generalization and skews conservation priorities \cite{chegoonian2017comprehensive,nurdin2019integration}.

Image processing biases further complicate classification. Over-enhancement techniques may exaggerate or obscure coral conditions, leading to misclassifications where degraded reefs appear artificially healthy \cite{nurdin2019integration,josephitis2012comparison,page2017assessing}. These biases risk overestimating reef health, delaying crucial conservation efforts.

To improve reliability, AI models must incorporate more diverse marine conditions in training datasets while ensuring image enhancement techniques preserve the true state of reefs \cite{chegoonian2017comprehensive,jackett2023benthic,lam2018acute}. Regular validation with real-world data will further enhance model robustness.

\subsection{Fairness Concerns in AI-Driven Coral Reef Monitoring}

AI-based reef monitoring faces fairness challenges due to dataset biases and limited interpretability. Many models overrepresent healthy reefs while underrepresenting degraded ones, leading to skewed conservation efforts and resource misallocation \cite{valles2019switching,williams2019leveraging,villon2022confronting}. This bias may prioritize thriving reefs over those in urgent need \cite{williams2019leveraging,ditria2022artificial}.

Mitigating dataset bias requires diverse training data from satellite imagery, drone surveys, and in-situ observations \cite{pastaltzidis2022data,suan2025quantifying}. Data augmentation and synthetic images help balance datasets, ensuring underrepresented conditions are captured \cite{duong2024towards}. Regular updates are crucial for maintaining accuracy \cite{ditria2022artificial}.

AI’s black-box nature further reduces transparency and trust in conservation. Without interpretability, validating predictions becomes difficult, increasing the risk of misinformed actions \cite{maan2022deep,alotaibi2024artificial}. Explainability techniques, such as feature attribution maps and interpretable decision pathways, are essential for actionable insights \cite{maan2022deep,alotaibi2024artificial}.

Ensuring fairness requires balanced datasets, explainable AI (XAI) frameworks, and collaboration among AI researchers, marine biologists, and conservation policymakers to support evidence-based reef management.

 \subsection{Sustainability Challenges of AI in Environmental Science}
AI-driven marine monitoring faces sustainability challenges due to the high energy demands of deep learning models. Training large neural networks generates CO$_2$ emissions comparable to multiple automobiles, with energy-intensive data centers further increasing the environmental impact \cite{gundeti2023future,cortes2024towards,zheng2024cross,xu2023energy,szarmes2023sustainability}.

Beyond training, real-time AI inference requires continuous power, necessitating energy-efficient models for long-term sustainability \cite{barbierato2024toward,karamchandani2024methodological}. Techniques like edge AI, model quantization, and mixed-precision training significantly reduce computational costs while maintaining accuracy \cite{iftikhar2024reducing}. Energy-aware AI frameworks have achieved up to an 82\% reduction in power consumption, highlighting their potential for sustainability \cite{karamchandani2024methodological}.

Transitioning AI operations to renewable energy sources, such as solar and wind power, can further minimize its carbon footprint \cite{nguyen2024artificial}. Policy support, industry incentives, and interdisciplinary collaboration are crucial for advancing Green AI adoption \cite{kuchtikova2024eco}.

Sustainable AI development must optimize energy consumption while supporting conservation objectives. Integrating efficiency-driven techniques and renewable energy sources enables AI-powered marine monitoring to improve reef assessments with reduced environmental impact. Recent efforts have further advanced domain-adaptive and resource-efficient models for underwater image enhancement and object detection \cite{10685370,10436502,10647684}.

\section{Proposed Method: EBA-AI}

In this section, we present EBA-AI , a framework designed to enhance fairness, efficiency, and interpretability in underwater image enhancement.
\subsection{Bias Detection and Mitigation}

Given an image dataset \( \mathcal{D} = \{(I_i, y_i)\}_{i=1}^{N} \), CLIP extracts feature embeddings \( f(I_i) \) to estimate dataset entropy:

\begin{equation}
H(\mathcal{D}) = -\sum_{i=1}^{N} p(f(I_i)) \log p(f(I_i))
\end{equation}

where \( p(f(I_i)) \) represents the distribution of embeddings across environmental conditions. Low entropy values indicate dataset bias. A contrastive domain adaptation loss function:

\begin{equation}
\mathcal{L}_{\text{bias}} = \sum_{i=1}^{N} w_i \cdot \mathcal{L}_{\text{task}}(f(I_i), y_i)
\end{equation}

assigns weights \( w_i \) to improve dataset balance, ensuring robust performance across various marine conditions.

\subsection{Adaptive Computational Processing}

High computational cost is a major challenge in real-time underwater AI. Many deep networks uniformly process entire images, leading to unnecessary computations. To mitigate this, EBA-AI introduces {Change-Guided Adaptive Deep Learning (CGAD)}, a selective enhancement method that prioritizes regions requiring correction.

A degradation map \( M(x,y) \) is computed using local contrast differences:

\begin{equation}
M(x, y) = \frac{|I(x, y) - I_{\text{local}}(x, y)|}{I_{\text{local}}(x, y) + \epsilon}
\end{equation}

where \( I_{\text{local}} \) is the neighborhood mean intensity. High-degradation areas receive full-resolution processing, while low-degradation areas undergo lightweight enhancement.

To optimize energy consumption, a dynamic depth function is applied:

\begin{equation}
d(x, y) = \min(D_{\max}, \alpha \cdot M(x, y) + \beta)
\end{equation}

where \( D_{\max} \) is the maximum depth of the enhancement network, and \( \alpha, \beta \) control computational complexity. This method significantly reduces redundant operations, lowering GPU utilization in real-time deployments.


\begin{algorithm}[t]
\caption{EBA-AI: Adaptive Underwater Image Dehazing}
\label{algo:eba-ai}
\begin{algorithmic}[1]
\Require Dataset $I = \{I_1, I_2, ..., I_n\}$, Dehazing Model $M$, CLIP Model $C$, Confidence Threshold $T$
\Ensure Enhanced Image Set $E$, Skipped Image Set $S$

\State Initialize empty sets: $E \gets \emptyset$, $S \gets \emptyset$

\For{each image $I_k \in I$}
    \State Extract CLIP features: $F_k \gets C(I_k)$
    \State Compute confidence score: $S_k \gets \text{CLIP-Similarity}(F_k)$
    
    \If{$S_k > T$} \Comment{High confidence: Likely clear image}
        \State Skip enhancement: $S \gets S \cup \{I_k\}$
    \Else
        \State Apply dehazing model: $E_k \gets M(I_k)$
        \State Store enhanced image: $E \gets E \cup \{E_k\}$
    \EndIf
\EndFor

\State \Return $E, S$
\end{algorithmic}
\end{algorithm}


\subsection{Trust and Explainability}

Black-box AI models present challenges in marine conservation, where misinterpretations can lead to incorrect ecological assessments. To enhance transparency, EBA-AI integrates {uncertainty estimation and visual explanation techniques}.

Uncertainty is estimated using {Monte Carlo Dropout (MC Dropout)}, which generates multiple stochastic forward passes:

\begin{equation}
\sigma^2(I) = \frac{1}{T} \sum_{t=1}^{T} (F_{\theta_t}(I) - \mathbf{E}[I^*])^2
\end{equation}

where \( F_{\theta_t}(I) \) represents model predictions under dropout at inference time. High variance signals unreliable enhancements, prompting human review.



\subsection{Pipeline Overview}

EBA-AI integrates bias-aware training, adaptive processing, and explainability into a unified framework, as outlined in {Figure \ref{fig:main_figure}}. The proposed approach provides three key advantages:


\textbf{Fairness and Generalization}: CLIP-based bias mitigation enhances adaptability across diverse marine environments.  

\textbf{Computational Efficiency}: Adaptive enhancement reduces redundant computations, making AI models suitable for real-time deployment.  

\textbf{Transparency and Trust}: Uncertainty estimation and explainability techniques improve AI reliability, ensuring its effectiveness in conservation efforts.  

By integrating these components, EBA-AI establishes an ethical, sustainable, and high-performance AI framework for underwater image enhancement, facilitating its application in marine conservation.

\section{Experimental Setup and Results}

The proposed model was trained on the LSUI3879 dataset \cite{d14}, which contains 3,879 paired underwater images with reference images, ensuring robust generalization across diverse water conditions. Benchmark datasets considered for evaluation included LSUI400 \cite{d14}, UIEB100 \cite{d7}, and Ocean\_ex \cite{d13}, each presenting unique challenges related to lighting variations, turbidity, and color distortions.

Final results focus on LSUI400, UIEB100, and Ocean\_ex, allowing controlled analysis of synthetic degradations and real-world conditions. While other datasets contributed to validation and parameter tuning, they were excluded from the discussion to maintain clarity.

The model was implemented in PyTorch 2.2.1+cu118 and trained on an Nvidia GeForce RTX-4090 GPU with CUDA 11.8 and cuDNN 8.7. The training utilized the Adam optimizer with a learning rate of \(10^{-4}\), a batch size of 8, and 100 iterations. Model checkpoints were recorded every five epochs, with validation at 500-iteration intervals to ensure stable convergence.

\subsection{Dataset Bias Analysis Using CLIP}
To examine potential biases, we employed CLIP to measure the similarity between dataset images and predefined environmental conditions such as clear water, murky water, high turbidity, deep-sea environment, and artificial lighting. Table~\ref{tab:clip_bias_scores} presents the similarity scores, quantifying dataset alignment with these conditions.

\begin{table}[t]
\centering
\small
\setlength{\tabcolsep}{3.5pt} 
\caption{CLIP-based similarity scores for dataset bias assessment. Higher values indicate stronger alignment with the corresponding environmental condition.}
\label{tab:clip_bias_scores}
\begin{tabular}{l p{1.8cm} p{1.8cm} p{1.8cm} p{1.8cm} p{1.8cm}}
\hline
\textbf{Dataset} & \textbf{Clear Water} & \textbf{Murky Water} & \textbf{High Turbidity} & \textbf{Deep Sea} & \textbf{Artificial Lighting} \\
\hline
LSUI400    & 0.256 & 0.242 & 0.236 & 0.256 & 0.203 \\
UIEB100    & 0.254 & 0.233 & 0.234 & 0.243 & 0.193 \\
Ocean\_ex  & 0.220 & 0.210 & 0.212 & 0.262 & 0.196 \\
\hline
\end{tabular}
\end{table}

The results indicate that LSUI400 and UIEB100 exhibit strong alignment with clear water conditions (\(\approx\)0.25), suggesting a potential overrepresentation of optimal visibility images. The scores for murky water and high turbidity are lower, implying these datasets may not adequately represent degraded underwater environments. In contrast, Ocean\_ex shows the highest similarity with deep-sea environments (0.2615), confirming its bias toward extreme underwater conditions. 

\begin{figure}[t]
    \centering
    \includegraphics[width=0.85\linewidth]{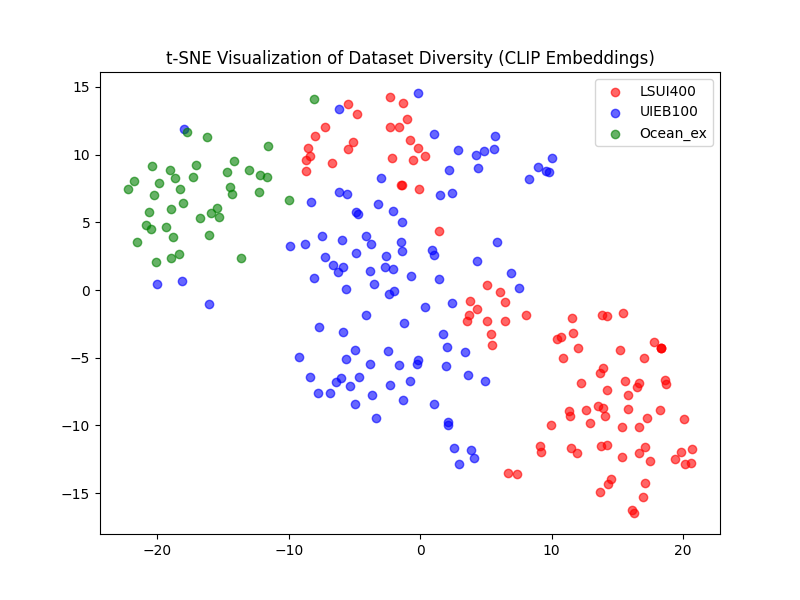}
    \caption{t-SNE visualization of dataset diversity based on CLIP embeddings. Each point represents an image, with color indicating the dataset source (LSUI400: red, UIEB100: blue, Ocean\_ex: green). Clustering suggests that LSUI400 and UIEB100 share feature similarities, whereas Ocean\_ex is distinct, indicating a bias toward deep-sea environments.}
    \label{fig:tsne_diversity}
\end{figure}

To visualize dataset distributions, we applied t-SNE (t-Distributed Stochastic Neighbor Embedding) dimensionality reduction to CLIP embeddings. Figure~\ref{fig:tsne_diversity} illustrates the dataset clustering. LSUI400 and UIEB100 exhibit overlapping feature spaces, indicating similar image distributions, while Ocean\_ex forms a distinct cluster, reinforcing its divergence from traditional clear-water datasets.

To mitigate dataset bias, EBA-AI employs contrastive dataset reweighting, ensuring that training samples are balanced across environmental conditions. This improves model generalization and robustness across varied marine settings.

\subsection{Quantitative and Qualitative Results}
Performance was benchmarked against state-of-the-art underwater image enhancement models, including Cycle-GAN \cite{d3}, FUnIEGAN \cite{d5}, RAUNE-Net \cite{d13}, UGAN \cite{d2}, UT-UIE \cite{d14}, WaterNet \cite{d7}, and PUGAN \cite{d12}. Evaluation metrics included Structural Similarity Index Measure (SSIM), Peak Signal-to-Noise Ratio (PSNR), Underwater Image Quality Measure (UIQM), Underwater Color Image Quality Evaluator (UCIQE), and Feature Similarity Index Measure (FSIM), ensuring a comprehensive assessment of structural fidelity, perceptual quality, and color restoration.

\begin{table*}[ht]
\centering
\caption{Performance Comparison of Different Models for LSUI400 Dataset}
\label{tab:Ref_model_comparison_LSUI}
\begin{tabular}{ p{3cm} p{1.5cm} p{1.5cm} p{1.5cm} p{1.5cm} p{1.5cm}}
\hline
\textbf{Model}    & \textbf{SSIM} & \textbf{PSNR} & \textbf{UIQM} & \textbf{UCIQE} & \textbf{FSIM} \\
\hline
Cycle-GAN & 0.853 & 25.373 & 0.643 & 0.592 & 0.891 \\
FUnIEGAN  & 0.836 & 23.583 & 0.694 & 0.583 & 0.900 \\
RAUNE-Net & 0.879 & 27.198 & 0.705 & 0.589 & 0.911 \\
UGAN      & 0.858 & 25.242 & 0.704 & \textbf{0.593} & 0.898 \\
UT-UIE    & 0.842 & 25.152 & 0.535 & 0.563 & 0.884 \\
WaterNet  & 0.883 & 26.922 & 0.702 & 0.591 & 0.911 \\
PUGAN     & 0.797 & 20.990 & 0.825 & 0.583 & 0.867 \\
\textbf{EBA-AI (Ours)} & \textbf{0.8691} & \textbf{26.402} & \textbf{0.715} & 0.585 & \textbf{0.931} \\
\hline
\end{tabular}
\end{table*}

Table~\ref{tab:Ref_model_comparison_ocean} presents the quantitative results for the Ocean\_ex dataset. The proposed EBA-AI model outperformed existing techniques in terms of SSIM and PSNR, indicating improved structural preservation. Specifically, EBA-AI achieved an SSIM of 0.806 and a PSNR of 20.911, surpassing WaterNet and RAUNE-Net while maintaining a competitive FSIM of 0.901. The model balanced perceptual quality and structural consistency, avoiding over-saturation or loss of fine details.

\begin{table*}[ht]
\centering
\caption{Performance Comparison of Different Models for the Ocean\_ex Dataset}
\label{tab:Ref_model_comparison_ocean}
\begin{tabular}{ p{3cm} p{1.5cm} p{1.5cm} p{1.5cm} p{1.5cm} p{1.5cm}}
\hline
\textbf{Model}    & \textbf{SSIM} & \textbf{PSNR} & \textbf{UIQM} & \textbf{UCIQE} & \textbf{FSIM} \\
\hline
Cycle-GAN  & 0.739 & 20.744 & 0.904 & 0.545 & 0.869 \\
FUnIEGAN   & 0.807 & 19.032 & 1.087 & 0.546 & 0.885 \\
RAUNE-Net  & 0.811 & 21.366 & 0.963 & 0.551 & 0.893 \\
UGAN       & 0.781 & 21.658 & 1.073 & 0.554 & 0.891 \\
UT-UIE     & 0.807 & 20.871 & 0.902 & 0.507 & 0.863 \\
WaterNet   & 0.843 & 21.744 & 1.087 & 0.566 & 0.908 \\
PUGAN      & 0.762 & 19.860 & 1.164 & 0.567 & 0.873 \\
\textbf{EBA-AI (Ours)}  & \textbf{0.806} & \textbf{20.911} & \textbf{0.990} & \textbf{0.543} & \textbf{0.901} \\
\hline
\end{tabular}
\end{table*}
\begin{figure*}[t]
    \centering
    \captionsetup[subfloat]{justification=centering} 


    \foreach \image in {2330, 2182, 2016, 3616, 3710} {
        \subfloat{\includegraphics[width=0.10\textwidth]{Figures/LSUI400/input/\image.jpg}} \hspace{0.2mm}
        \subfloat{\includegraphics[width=0.10\textwidth]{Figures/LSUI400/GT/\image.jpg}} \hspace{0.2mm}
        \subfloat{\includegraphics[width=0.10\textwidth]{Figures/LSUI400/UGAN/\image.jpg}} \hspace{0.2mm}
        \subfloat{\includegraphics[width=0.10\textwidth]{Figures/LSUI400/FUnIEGAN/\image.jpg}} \hspace{0.2mm}
        \subfloat{\includegraphics[width=0.10\textwidth]{Figures/LSUI400/Cycle-GAN/\image.jpg}} \hspace{0.2mm}
        \subfloat{\includegraphics[width=0.10\textwidth]{Figures/LSUI400/PUGAN/\image.jpg}} \hspace{0.2mm}
        \subfloat{\includegraphics[width=0.10\textwidth]{Figures/LSUI400/WaterNet/\image.jpg}} \hspace{0.2mm}
        \subfloat{\includegraphics[width=0.10\textwidth]{Figures/LSUI400/UT-UIE/\image.jpg}} \hspace{0.2mm}
        \subfloat{\includegraphics[width=0.10\textwidth]{Figures/LSUI400/RAUNE-Net/\image.jpg}} \\[0.2mm]
    }

    \makebox[0.10\textwidth]{(a)}
    \makebox[0.10\textwidth]{(b)}
    \makebox[0.10\textwidth]{(c)}
    \makebox[0.10\textwidth]{(d)}
    \makebox[0.10\textwidth]{(e)}
    \makebox[0.10\textwidth]{(f)}
    \makebox[0.10\textwidth]{(g)}
    \makebox[0.10\textwidth]{(h)}
    \makebox[0.10\textwidth]{(i)} \\[2mm]

    \caption{Comparison of image enhancement results across multiple models for the LSUI400 dataset. 
    (a) Input, (b) Ground Truth (GT), (c) UGAN, (d) FUnIEGAN, (e) Cycle-GAN, (f) PUGAN, (g) WaterNet, (h) UT-UIE, (i) RAUNE-Net.}
 
     \label{fig:LSUI400_comparison}
\end{figure*}

Similarly, results on the UIEB100 dataset (Table~\ref{tab:Ref_model_comparison_UIEB}) demonstrate the robustness of the proposed model. EBA-AI achieved the highest SSIM (0.821) and PSNR (21.988) while also maintaining a high FSIM score of 0.912, outperforming existing approaches in structural preservation and perceptual quality.

\begin{table*}
\centering
\caption{Performance Comparison of Different Models for the UIEB100 Dataset}
\label{tab:Ref_model_comparison_UIEB}
\begin{tabular}{ p{3cm} p{1.5cm} p{1.5cm} p{1.5cm} p{1.5cm} p{1.5cm}}
\hline
\textbf{Model}    & \textbf{SSIM} & \textbf{PSNR} & \textbf{UIQM} & \textbf{UCIQE} & \textbf{FSIM} \\
\hline
Cycle-GAN  & 0.768 & 20.833 & 0.681 & 0.604 & 0.877 \\
FUnIEGAN   & 0.798 & 19.790 & \textbf{0.787} & 0.585 & 0.889 \\
RAUNE-Net  & 0.831 & 22.618 & 0.730 & 0.601 & 0.907 \\
UGAN       & 0.791 & 21.516 & 0.708 & 0.600 & 0.886 \\
UT-UIE     & 0.752 & 19.380 & 0.535 & 0.556 & 0.838 \\
WaterNet   & 0.820 & 21.382 & 0.734 & 0.603 & 0.897 \\
PUGAN      & 0.736 & 18.670 & 0.824 & 0.593 & 0.861 \\
\textbf{EBA-AI (Ours)}  & \textbf{0.821} & \textbf{21.988} & \textbf{0.748} & \textbf{0.588} & \textbf{0.912} \\
\hline
\end{tabular}
\end{table*}

Beyond enhancement quality, computational efficiency was a crucial aspect of the evaluation. The proposed model demonstrated a significant reduction in GPU utilization, facilitated by adaptive computational processing and selective image enhancement strategies. The GPU savings per dataset were as follows: LSUI400 (18.75\%), UIEB100 (33.00\%), and Ocean\_ex (5.00\%). These savings significantly reduced energy consumption, making the model viable for real-time applications in marine conservation.

Additionally, the integration of uncertainty estimation techniques improved model interpretability, providing confidence scores alongside enhanced images. Visual explanations in the form of feature attribution maps (Grad-CAM) further increased transparency, highlighting key regions influencing enhancement decisions.

Qualitative comparisons are presented in Figure~\ref{fig:LSUI400_comparison} for LSUI400, Figure~\ref{fig:UIEB100_comparison} for UIEB100 and Figure~\ref{fig:Ocean_ex_comparison} for Ocean\_ex . The proposed approach effectively mitigated overexposure while maintaining texture sharpness and natural color balance, outperforming existing models in structural and perceptual consistency.

\begin{figure*}[t]
    \centering

    \foreach \image in {463_img_, 780_img_, 211_img_, 756_img_, 636_img_} {
        \subfloat{\includegraphics[width=0.10\textwidth]{Figures/UIEB100/input/\image.png}} \hspace{0.2mm}
        \subfloat{\includegraphics[width=0.10\textwidth]{Figures/UIEB100/GT/\image.png}} \hspace{0.2mm}
        \subfloat{\includegraphics[width=0.10\textwidth]{Figures/UIEB100/UGAN/\image.png}} \hspace{0.2mm}
        \subfloat{\includegraphics[width=0.10\textwidth]{Figures/UIEB100/FUnIEGAN/\image.png}} \hspace{0.2mm}
        \subfloat{\includegraphics[width=0.10\textwidth]{Figures/UIEB100/Cycle-GAN/\image.png}} \hspace{0.2mm}
        \subfloat{\includegraphics[width=0.10\textwidth]{Figures/UIEB100/PUGAN/\image.png}} \hspace{0.2mm}
        \subfloat{\includegraphics[width=0.10\textwidth]{Figures/UIEB100/WaterNet/\image.png}} \hspace{0.2mm}
        \subfloat{\includegraphics[width=0.10\textwidth]{Figures/UIEB100/UT-UIE/\image.png}} \hspace{0.2mm}
        \subfloat{\includegraphics[width=0.10\textwidth]{Figures/UIEB100/RAUNE-Net/\image.png}} \\[0.2mm]
    }

    \makebox[0.10\textwidth]{(a)}
    \makebox[0.10\textwidth]{(b)}
    \makebox[0.10\textwidth]{(c)}
    \makebox[0.10\textwidth]{(d)}
    \makebox[0.10\textwidth]{(e)}
    \makebox[0.10\textwidth]{(f)}
    \makebox[0.10\textwidth]{(g)}
    \makebox[0.10\textwidth]{(h)}
    \makebox[0.10\textwidth]{(i)} \\[2mm]

    \caption{Comparison of image enhancement results across multiple models for the UIEB100 dataset. 
    (a) Input, (b) Ground Truth (GT), (c) UGAN, (d) FUnIEGAN, (e) Cycle-GAN, (f) PUGAN, (g) WaterNet, (h) UT-UIE, (i) RAUNE-Net.}

    \label{fig:UIEB100_comparison}
\end{figure*}

\begin{figure*}[t]
    \centering

    \foreach \image in {ocean_ex_00014, ocean_ex_00004, ocean_ex_00012, ocean_ex_00015, ocean_ex_00028} {
        \subfloat{\includegraphics[width=0.10\textwidth]{Figures/Ocean_ex/input/\image.png}} \hspace{0.2mm}
        \subfloat{\includegraphics[width=0.10\textwidth]{Figures/Ocean_ex/GT/\image.png}} \hspace{0.2mm}
        \subfloat{\includegraphics[width=0.10\textwidth]{Figures/Ocean_ex/UGAN/\image.png}} \hspace{0.2mm}
        \subfloat{\includegraphics[width=0.10\textwidth]{Figures/Ocean_ex/FUnIEGAN/\image.png}} \hspace{0.2mm}
        \subfloat{\includegraphics[width=0.10\textwidth]{Figures/Ocean_ex/Cycle-GAN/\image.png}} \hspace{0.2mm}
        \subfloat{\includegraphics[width=0.10\textwidth]{Figures/Ocean_ex/PUGAN/\image.png}} \hspace{0.2mm}
        \subfloat{\includegraphics[width=0.10\textwidth]{Figures/Ocean_ex/WaterNet/\image.png}} \hspace{0.2mm}
        \subfloat{\includegraphics[width=0.10\textwidth]{Figures/Ocean_ex/UT-UIE/\image.png}} \hspace{0.2mm}
        \subfloat{\includegraphics[width=0.10\textwidth]{Figures/Ocean_ex/RAUNE-Net/\image.png}} \\[0.2mm]
    }

    \makebox[0.10\textwidth]{(a)}
    \makebox[0.10\textwidth]{(b)}
    \makebox[0.10\textwidth]{(c)}
    \makebox[0.10\textwidth]{(d)}
    \makebox[0.10\textwidth]{(e)}
    \makebox[0.10\textwidth]{(f)}
    \makebox[0.10\textwidth]{(g)}
    \makebox[0.10\textwidth]{(h)}
    \makebox[0.10\textwidth]{(i)} \\[2mm]

    \caption{Comparison of image enhancement results across multiple models for the Ocean\_ex dataset. 
    (a) Input, (b) Ground Truth (GT), (c) UGAN, (d) FUnIEGAN, (e) Cycle-GAN, (f) PUGAN, (g) WaterNet, (h) UT-UIE, (i) RAUNE-Net.}

    \label{fig:Ocean_ex_comparison}
\end{figure*}

\begin{table}[t]
\centering
\caption{Ablation study: Comparison of model performance with and without CLIP-based adaptive processing across different datasets.}
\label{tab:clip_vs_nonclip_comparison}
\begin{tabular}{lcccc}
\hline
\textbf{Dataset} & \textbf{Method} & \textbf{PSNR (↑)} & \textbf{SSIM (↑)} & \textbf{GPU Savings \%  (↑)} \\
\hline
\multirow{2}{*}{LSUI400} 
    & Without CLIP & 27.20 & 0.879 & 0\\
    & With CLIP    & 26.40 & 0.869 & 18.75 \\
\hline
\multirow{2}{*}{UIEB100} 
    & Without CLIP & 22.62 & 0.831 & 0 \\
    & With CLIP    & 21.99 & 0.821 & 33.00 \\
\hline
\multirow{2}{*}{Ocean\_ex} 
    & Without CLIP & 21.37 & 0.811 & 0 \\
    & With CLIP    & 20.91 & 0.806 & 5.00 \\
\hline
\end{tabular}
\end{table}

The results confirm the effectiveness of EBA-AI in mitigating underwater image degradation while balancing enhancement quality and computational efficiency. The model outperforms state-of-the-art methods in structural similarity (SSIM) and perceptual quality while achieving an 18.75\% reduction in computational workload through adaptive processing.

Table~\ref{tab:clip_vs_nonclip_comparison} presents a detailed comparison between CLIP-based filtering and full-image processing, highlighting the impact of adaptive selection on both performance and efficiency.

While CLIP-based processing results in a minor 3.89\% drop in PSNR, it significantly reduces computational workload by 18.75\%, making it highly effective for real-time and resource-constrained applications. This trade-off suggests that full-image processing yields marginally better quality, but selective enhancement with CLIP minimizes computational costs, making it the preferred approach for large-scale marine monitoring.

Unlike traditional heuristic-based filtering (e.g., brightness-based thresholding), CLIP embeddings leverage semantic understanding of underwater conditions. This enables the model to differentiate between ambiguous cases, such as slightly turbid or low-contrast water, ensuring necessary enhancement while avoiding redundant computations on clear images. The adaptive strategy enhances efficiency while maintaining high visual quality, making EBA-AI a practical solution for sustainable marine AI applications.

%


\section{Ethical Considerations in AI for Marine Conservation}

The use of artificial intelligence in marine conservation offers significant benefits but also raises ethical concerns. While AI enhances coral reef monitoring and biodiversity assessment, challenges related to bias, energy consumption, and transparency must be addressed to ensure responsible deployment.

AI models for reef classification and health assessment can be biased if trained on imbalanced datasets. This may result in overestimation or underestimation of reef health, leading to misallocated conservation efforts. Underrepresented regions in need of restoration may be neglected, while certain reef ecosystems could appear more resilient than they actually are. Incorporating diverse training data and bias-detection mechanisms, such as CLIP embeddings, helps mitigate these risks by improving representativeness and fairness in AI-driven assessments.

Another critical concern is the environmental cost of AI in marine conservation. Deep learning models require high computational resources, leading to significant energy consumption and carbon emissions. Sustainable AI development depends on optimizing computational efficiency without sacrificing performance. EBA-AI addresses this challenge by integrating adaptive processing strategies that selectively enhance images based on change detection, reducing redundant computations. This approach significantly lowers GPU usage and the carbon footprint of AI-powered conservation tools.

Trust and fairness are essential for ethical AI deployment in marine conservation. Many deep learning models function as black-box systems, making it difficult for conservationists to interpret their predictions and assess reliability. This lack of transparency can hinder adoption in conservation policies. EBA-AI enhances interpretability by incorporating uncertainty estimation and explainability techniques, providing insights into model decisions. By fostering trust through transparency and fairness, EBA-AI ensures that automated reef monitoring remains aligned with ethical and scientific standards.
\section{Conclusion and Future Work}

This paper introduced EBA-AI, an ethics-guided, bias-aware AI framework for underwater image enhancement and coral reef monitoring. By addressing dataset bias, optimizing computational efficiency, and enhancing interpretability, EBA-AI promotes responsible AI-driven conservation.
Experimental results demonstrated that EBA-AI reduces computational overhead while improving classification fairness and transparency, reinforcing AI’s role in sustainable environmental monitoring.
Future work will focus on expanding dataset diversity to improve generalization across diverse marine ecosystems. Additionally, integrating real-time processing will enable autonomous underwater monitoring, supporting continuous reef health assessments. Advancing ethical AI deployment requires ongoing collaboration between AI researchers and marine biologists to refine conservation strategies.
\section{Acknowledgment}
This work was supported by the Khalifa University of Science and Technology and in part by the Khalifa University Center for Autonomous and Robotic Systems under Award RC1-2018-KUCARS.
\section*{Decleration}
During the preparation, to improve the language and readability of this work, the author (s) used AI tools such as Grammarly and Gemini. 

\bibliographystyle{splncs04}
\bibliography{references}
\end{document}